\documentclass{article}

\usepackage{arxiv}

\usepackage[utf8]{inputenc} 
\usepackage[T1]{fontenc}    
\usepackage{hyperref}       
\usepackage{url}            
\usepackage{booktabs}       
\usepackage{amsfonts}       
\usepackage{amsmath}
\usepackage{nicefrac}       
\usepackage{microtype}      
\usepackage{lipsum}
\usepackage{graphicx}
\usepackage{xcolor}
\usepackage{natbib}
\usepackage{textcomp, gensymb} 
\graphicspath{ {./Figs/} }

\title{Few TensoRF: Enhance the Few-shot on Tensorial Radiance Fields}

\author{
 Thanh-Hai Le \\
  School of Computer Science and Engineering\\
  The Saigon International University\\
  Ho Chi Minh City 700000, Vietnam\\
  \texttt{lethanhhai@siu.edu.vn} \\ 
  \And
 Hoang-Hau Tran \\
  Department of Informatic Technology\\
  FPT University\\
  Ho Chi Minh City 700000, Vietnam\\
  \texttt{hauth151239@fpt.edu.com} \\  
  \And
 Trong-Nghia Vu \\
  Department of Informatic Technology\\
  FPT University\\
  Ho Chi Minh City 700000, Vietnam\\
  \texttt{nghiavt150934@fpt.edu.com} \\    
}

\begin{document}
\maketitle
\begin{abstract}
This paper presents Few TensoRF, a 3D reconstruction framework that combines TensorRF's efficient tensor based representation with FreeNeRF's frequency driven few shot regularization. Using TensorRF to significantly accelerate rendering speed and introducing frequency and occlusion masks, the method improves stability and reconstruction quality under sparse input views. Experiments on the Synthesis NeRF benchmark show that Few TensoRF method improves the average PSNR from 21.45 dB (TensorRF) to 23.70 dB, with the fine tuned version reaching 24.52 dB, while maintaining TensorRF's fast \(\approx10-15\) minute training time. Experiments on the THuman 2.0 dataset further demonstrate competitive performance in human body reconstruction, achieving 27.37-34.00 dB with only eight input images. These results highlight Few TensoRF as an efficient and data effective solution for real-time 3D reconstruction across diverse scenes.
\end{abstract}


\section{Introduction}
\label{sec:intro}
The ability to create accurate and versatile 3D reconstructions has become increasingly vital in today’s technological landscape. From applications in computer vision to healthcare, and even in the world of entertainment, 3D reconstruction plays a pivotal role. The history of 3D reconstruction methods has witnessed significant advancements, with a continuous pursuit of techniques that can transform single or multiple 2D images into detailed 3D representations. This study, inspired by the success of the renowned Neural
Radiance Field (NeRF) \cite{mildenhall2020nerf, brualla2021nerf} for synthesizing novel views of complex scenes, aims to incorporate two improved methods related to NeRF: the Tensorial Radiance Field (TensorRF) \cite{chen2022tensorf} and the enhancement of few-shot neural rendering with Free Frequency Regularization (FreeNeRF) \cite{yang2023freenerf}. TensoRF introduces a novel approach using 4D tensors and decompositions \cite{caroll1970analysis, lathauwer2008decomposition}, for improved rendering quality, reduced memory usage, and faster reconstruction times compared to NeRF. On the other hand, FreeNeRF simplifies few-shot neural rendering by introducing costeffective regularization terms, outperforming complex alternatives with a single line of code change, and emphasizing the significance of frequency in NeRF’s training. This study is motivated by the pressing need for a 3D reconstruction method that is both versatile and efficient. Our primary objective is to develop a versatile 3D reconstruction method, leveraging the combined capabilities of TensorRF and FreeNeRF. This approach is designed to address the shortcomings of previous methods and promote broader applications in various domains, especially experience on human body datasets. In developing our approach, we build upon the foundation of TensoRF and introduce two primary regularization techniques, frequency, and occlusion, inspired by the FreeNeRF paper. These techniques are incorporated as a positional encoding step to map each input 5D coordinate within the tensor component of the scene. Our methodology will be thoroughly compared with previous NeRF-related methods, not only on typical objects but also on the more complex task of 3D human body reconstruction. The THuman2.0 dataset \cite{yu2021function4d} will be used and trained on different setups for experiments and comparison.

\section{Related works}
\label{sec:relatedworks}
\subsection{Neural Radiance Fields (NeRF)}
Neural Radiance Fields (NeRF) \cite{mildenhall2020nerf} has transformed the approach to 3D reconstruction. Utilizing fully connected deep networks, NeRF processes single continuous 5D coordinates, including spatial location \textit{(x, y, z)} and viewing direction \((\theta, \phi\)) to generate volume density and view-dependent emitted radiance. These properties facilitate the synthesis of novel views by querying 5D coordinates along camera rays and utilizing volume rendering techniques. NeRF’s applications range from novel view synthesis \cite{brualla2021nerf}, 3D generation \cite{poole2022dream, jain2022zeroshot} to deformation \cite{park2021nerfies, pumarola2020dnerf} and video \cite{li2020neuralscene, Xian2020spacetime}. Despite its remarkable capabilities in high-quality scene representation, NeRF’s significant drawback is its demand for a substantial number of input images, rendering it ineffective at synthesizing novel views from a limited set of input views, such as 3, 6, or 9 views \cite{yang2023freenerf}. This limitation not only hinders its potential real-world applications but also contributes to its relatively slow rendering speed, with NeRF requiring approximately 35 hours of training for completeness \cite{chen2022tensorf}.

\subsection{Tensorial Radiance Field (TensorRF)}
Tensorial Radiance Fields (TensorRF) \cite{chen2022tensorf} presents a novel approach to 3D reconstruction by representing the radiance field as a 4D tensor. TensorRF utilizes the tensor decomposition technique \cite{Aaneas2016largescale} to achieve improved rendering quality, reduced memory usage, and faster reconstruction times compared to NeRF. Although the experience and comparison between the classic CP decomposition \cite{kolda2009tensor} and a new vector matrix (VM) decomposition, TensorRF focuses mainly on the reconstruction through gradient descent rather than the decomposition itself. This introduces low-rank regularization in the optimization process, ultimately leading to enhanced rendering quality. TensorRF’s advantage lies in its potential to make 3D reconstruction more memory-efficient while maintaining rendering quality.

\subsection{FreeNeRF: Few-shot Neural Rendering}
Few-shot neural rendering has posed a challenge, often requiring complex external information or pretraining. FreeNeRF \cite{yang2023freenerf} addresses this challenge by introducing cost-effective regularization terms, notably frequency and occlusion regularization. Remarkably, these modifications can be incorporated with minimal changes to the original NeRF code, maintaining the same computational efficiency. This approach underscores the importance of geometry in few-shot neural rendering, achieving impressive results without expensive pre-training or complex training-time patch rendering.

\subsection{Human Body 3D Reconstruction}
Reconstructing the 3D Human Body is the focus of our latest experiment, aimed at expanding the capabilities of NeRF-like methods beyond their traditional applications on standard datasets like Blender, DTU \cite{Aaneas2016largescale, jensen2014largescale}, or LLFF \cite{mildenhall2019locallight}. Unlike the objects found in these datasets, the challenges posed by a comprehensive Human Body dataset such as THuman 2.0 \cite{yu2021function4d} are distinct, ranging from the diverse human shapes and clothing variations to the multitude of poses encountered in real-world scenarios. Our innovative approach, rooted in NeRF-like principles, prioritize generating novel views over constructing the 3D mesh or voxel itself— a departure from earlier methods like ICON \cite{xiu2022icon}, ECON \cite{xiu2023econ} and others. This unique perspective holds promise, particularly in scenarios with limited resources, offering a viable solution to address the complexities inherent in capturing the richness of the human form.

\section{Method}
\label{sec:method}
\begin{figure}[htb!] 
    \centering
    \includegraphics[width=0.8\textwidth]{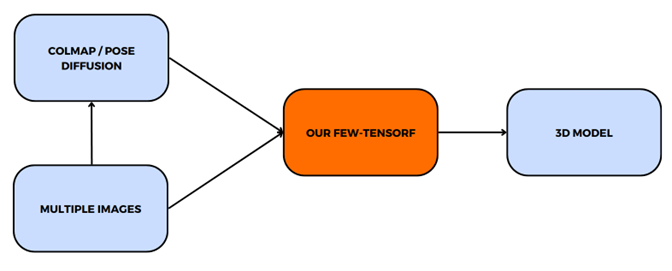}
    \caption{\textbf{Few TensoRF in the 3D NeRF-like Pipeline.} The diagram shows where Few TensoRF fits into the core of the 3D NeRF-like reconstruction pipeline. It all begins with multiple scene images, and either COLMAP or Pose Diffusion calculates precise camera details. These ones, along with the images, train Few TensoRF. The trained Few TensoRF model then takes center stage in the final steps, ensuring a streamlined and accurate 3D model reconstruction within the NeRF-like framework. This visual guide underscores the seamless integration of Few TensoRF at the heart of the 3D NeRF-like process.}
    \label{fig:fig1}
\end{figure}

The proposed Few-TensoRF is still a 3D reconstruction method, capable of exporting 3D object meshes due to its special 3D point coordinate and RGB with density mapping. However, it functions as just one core process within the broader NeRF application-level pipeline, as depicted in Fig. \ref{fig:fig1}.
The Few TensoRF method mainly takes advantage of the strong point in 2 methods: faster training time with TensoRF \cite{chen2022tensorf} and improvement of the FreeNeRF \cite{yang2023freenerf} in sparse input cases.

\subsection{Fast training time with Tensorial Radiance Field (TensoRF) Base}
\begin{figure}[htb!] 
    \centering
    \includegraphics[width=1\textwidth]{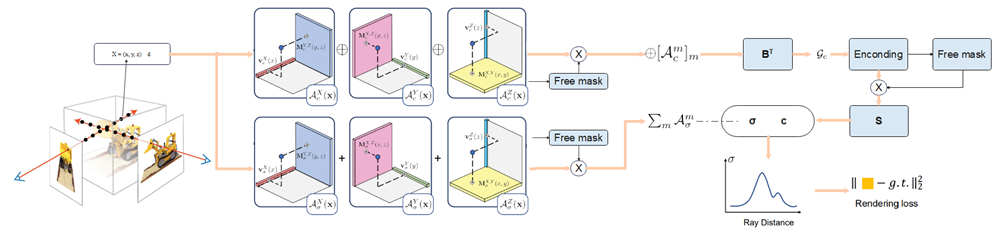}
    \caption{\textbf{The Few TensoRF overview.} This illustration was adapted from the original TensoRF paper \cite{chen2022tensorf}. It highlights two significant enhancements as we move from left to right: incorporating density frequency masks and integrating appearance color frequency on each rendered voxel. Additionally, a supplementary frequency mask is applied to refine the positional encoding fed into the neural networks.}
    \label{fig:fig2}
\end{figure}

TensoRF builds upon the NeRF framework, introducing crucial improvements in the form of a neural network architecture. While both TensoRF and NeRF share the overarching goal of mapping any 3D location \(x\) and viewing direction \(d\) to their respective volume density \(\sigma\) and, view-dependent color \(c\), TensoRF distinguishes itself by adopting a unique approach. In contrast to NeRF, which relies solely on Multi-Layer Perceptrons (MLPs), TensoRF models the radiance field of a scene as a 4D tensor. This innovative methodology uses a regular 3D grid \(G\) with per-voxel multi-channel features to represent the scene’s radiance field. In simple terms, TensoRF separately models the volume density \(\sigma\) and view-dependent color \(c\) by two distinct grids: a geometry grid \(G\) and an appearance grid \(G_c\). The continuous grid-based radiance field is expressed as:

\begin{equation}\label{eqn:eq1}
   {\sigma, c} = G_\sigma(x),S(G_c(x),d)
\end{equation}
Where \(G(x)\) and \(G_c(x)\) represent the trilinear interpolated features from the two grids at location \(x\). Both \(G(x)\) and \(G_c(x)\) are modeled as factorized tensors, employing the VM decomposition method. This relationship can be expressed using the following formulas (or as shown in Fig. \ref{fig:fig2}):

\begin{equation}\label{eqn:eq2}
   {G_\sigma(x)} = \sum_{r=1}^{R_\sigma} \sum_{m \in XYZ}A_{\sigma,r}^{m}
\end{equation}

\begin{equation}\label{eqn:eq3}
   {G_c(x)} = \sum_{r=1}^{R_o}{A_{c,r}^{X} \circ b_{3r-2} + A_{c,r}^{Y} \circ b_{3r-2} + A_{c,r}^{Z} \circ b_{3r}}
\end{equation}

As detailed in the original TensoRF paper \cite{chen2022tensorf}, the function \(S\) (in Eqn. \ref{eqn:eq1}) is introduced as a pre-selected function. This function, which can be a small MLP or a spherical harmonics (SH) function, is employed to convert an appearance feature vector and a viewing direction \(d\) into the final color \(c\). However, in this study, we will take advantage of the variant of MLP structures to get more customization for few-shot scenarios. Those \(A\) and \(A_c\) tensor components play a pivotal role in the revolution of our Few TensorRF technique.

\subsection{Enhance Few-shot on Tensorial Radiance Field (Few TensoRF)}
\begin{figure}[htb!] 
    \centering
    \includegraphics[width=0.8\textwidth]{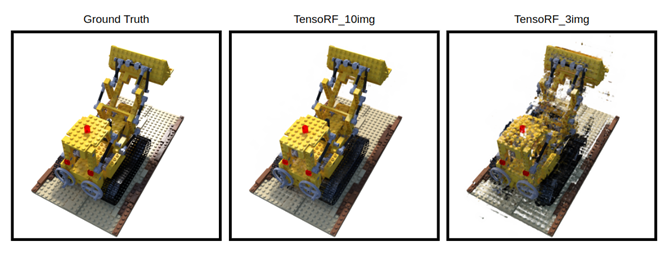}
    \caption{\textbf{Comparing the rendered novel views quality of TensoRF baseline across different setups.} To assess TensoRF’s performance on sparse input scenarios, we retrain the default TensoRF setup on different amounts of training image inputs. Following the image order, we have the original testing image (Ground Truth), and the novel view images rendered with TensoRF after training with 10 and 3 images (TensoRF\_10img and TensorRF\_10img) respectively.}
    \label{fig:fig3}
\end{figure}

Given the findings depicted in Figure \ref{fig:fig3}, showcasing the rendered novel views with varying amounts of training images in the TensoRF baseline, it becomes evident that TensoRF encounters challenges in delivering satisfactory results under sparse input conditions. The visual comparison highlights a noticeable disparity in the quality of novel views, particularly as the number of training images decreases. As discussed in the FreeNeRF \cite{yang2023freenerf}, we anticipate encountering issues arising from over-fast convergence during the optimization process. This rapid convergence, particularly in high-frequency components, hinders TensorRF’s ability to effectively explore low frequency information. Consequently, it introduces a quite bias in TensoRF, leading to the manifestation of undesired high-frequency artifacts (TensoRF 3img in Fig. \ref{fig:fig3}).

Therefore, to achieve better results in sparse input cases, the proposed Few TensoRF method proposes three improvements on the TensoRF \cite{chen2022tensorf} baseline, as depicted in Figure \ref{fig:fig2}. Frequency Masking Tensor Components. The goal here is to decrease the sensitivity of each tensor component in the high-frequency domain during the initial stages of training. This approach guides our model to concentrate on enhancing low-frequency structures, ensuring the stability of tensor components against undesired high-frequency artifacts. As mentioned earlier, there are two types of components: A for the density component and Ac for the appearance color feature component. Two frequency masks will be applied to them according to the following equations:

\begin{equation}\label{eqn:eq4}
   {A_{L}^{'}(t,T;\mathbf{x})} = {A_{L}(\mathbf{x}) \odot \alpha(t,T,L)}
\end{equation}

with 
\begin{equation*}\label{eqn:eq5}
   {\alpha_{i}(t,T,L)} = \left\{\begin{array}{lcl} 
   1 & \mbox{if} & {i\leq\frac{t \cdot L}{T}}\\ 
   \frac{t \cdot L}{T} - \lfloor{\frac{t \cdot L}{T}}\rfloor & \mbox{if} & \frac{t \cdot L}{T} + 3 < i \leq \frac{t \cdot L}{T} + 6 \\
   0 & \mbox{if} & i > \frac{t \cdot L}{T} + 6
   \end{array}\right.
\end{equation*}

Where \(\alpha_i(t,T,L)\) denotes the \(i^{th}\) bit value of \(\alpha(t,T,L)\) ; \(t\) and \(T\) are the current training iteration and the total iterations, respectively, \(L\) corresponds to either the density component length or the appearance feature length, depending on the desired tensor component to mask. This dynamic frequency mask, a unique feature introduced in FreeNeRF \cite{yang2023freenerf}, calculates the mask values based on the current iteration if \(t\) is less than \(T\). The calculation involves both an integer and a fractional part, ensuring a smooth transition. If \(t\) is greater than or equal to \(T\), the method returns a tensor of ones. This dynamic frequency mask is particularly beneficial for enhancing the high-frequency details of rendering novel views in the later stages of training. Additionally, we have implemented a simplified version of Eqn. \ref{eqn:eq4}, allowing us to set a fixed visible ratio for the frequency mask. This can be expressed as follows: 

\begin{equation}\label{eqn:eq6}
   {Freq\_mask[:int(L*v\_ratio)]} = 1
\end{equation}

\begin{figure}[htb!] 
    \centering
    \includegraphics[width=1.0\textwidth]{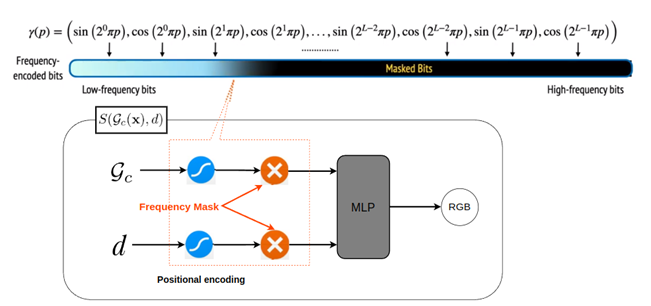}
    \caption{\textbf{Diagram illustrating Frequency Masking on the appearance grid \(G_c\) and view direction \(d\).}}
    \label{fig:fig4}
\end{figure}

\textit{Frequency Masking Appearance Grid \(G_c\)}. Similar to the process of masking tensor components, we once again apply the frequency mask, this time to the entire appearance grid \(G_c\), along with the viewing direction \(d\) as described in Fig. \ref{fig:fig4}. The masking technique employed here is analogous to the frequency masking tensor component, intended to mitigate overfitting in the MLP neural network associated with high-frequency signals. This frequency regularization acts as a filter for the positional encoding of the appearance grid \(G_c\) and view direction \(d\) (akin to Eqn. \ref{eqn:eq4}). Subsequently, the output is fed into a streamlined neural network, or MLP, responsible for predicting the RGB value of the input point.

\textit{Occlusion Regularization}. Despite efforts like frequency regularization, characteristic artifacts, such as ”walls” or ”floaters,” persist in novel views during few-shot neural rendering. Therefore, the novel ”occlusion” regularization is introduced to push the density of floaters in the near-camera region to zeros, and models will learn to explain this area in a farther place. This occlusion loss is directly incorporated into the original TensoRF-based loss, originally designed for overfitting but now tailored to be more adaptable in few-shot scenarios, reducing overfitting tendencies. This integration aims to enhance the model’s robustness, providing a more balanced performance in handling sparse input cases.

\section{Results and Discussion}
\label{sec:results}
\subsection{Experiment setup}
\textit{Dataset and Metric}. We utilized two datasets with distinct purposes: Synthesis NeRF and THuman 2.0. For performance evaluation, we replicate the few-shot Synthesis NeRF dataset following the methodology outlined in FreeNeRF. THuman 2.0 is utilized by the Blender Nerf add-on to extract 2D images and corresponding camera information, primarily serving new experiments. We will use PSNR as the core metric to evaluate the qualitative performance of the methods.

\textit{Implementation}. To ensure fair comparisons among different methods, consistent experimental environments were maintained for each training session. All experiment methods were implemented or reproduced with a PyTorch-based framework. The four main methods explored in this study include the reproduced few-shot state-of-the-art FreeNeRF, the original TensoRF on a few-shot dataset setup, and two variations of the proposed Few TensoRF, facilitating performance comparisons with the former two. Fundamentally, our models were trained with a ray batch size of 1024 over 15,000 iterations. Optimization was performed using the Adam optimizer with a specific learning rate, and training was conducted on Google Colab with a T4 Tesla GPU. 

\subsection{Comparison}
\textit{Synthesis NeRF Comparison}. As shown in Table \ref{tab:table1}, the Our Few TensorRF demonstrates significant improvements in average PSNR compared to both FreeNeRF (50k iters, repro.) and TensorRF (repro.). The finetuned version of Our Few TensorRF further enhances performance, achieving the highest PSNR in several scenes. Notably, the raw version of Our Few TensorRF consistently outperforms TensoRF (repo.), underscoring its efficacy in few-shot scenarios and showcasing advancements introduced by Our Few TensorRF. Fine-tuning Our Few TensorRF leads to additional improvements, surpassing both FreeNeRF and TensorRF in almost all scenes. These results underscore the effectiveness of our proposed Few TensorRF methods in achieving superior rendering quality, particularly in challenging few-shot conditions. However, there is a noticeable challenge in the Drums scene. This anomaly may be attributed to the intricate details and complexity of the Drum scene, making it the most challenging in the Synthesis NeRF dataset. In scenarios with numerous hidden angles, our model may struggle to comprehend the intricacies, leading to the reconstruction of some peculiar artifacts.

\begin{table}[htb!]
 \caption{PSNR metrics comparison on each Synthesis NeRF’s scene. The best-performing and second-best results from our experiments are denoted by red and blue highlights, respectively.}
  \centering
  \begin{tabular}{llllllllll}
    \toprule    
    \textbf{PSNR} & \textbf{Lego} & \textbf{Chair} & \textbf{Drums} & \textbf{Ficus} & \textbf{Mic} & \textbf{Ship} & \textbf{Materials} & \textbf{Hotdog} & \textbf{Avg.}\\    
    \midrule
    FreeNeRF\_50k iters (repro.) & 24.81  & 27.01 & \textcolor{red}{20.46} & 20.59 & \textcolor{blue}{25.54} & \textcolor{red}{23.79} & 21.69 & \textcolor{blue}{29.40} & \textcolor{blue}{24.16} \\
    Tensorf (repro.) & 23.83 & 23.45 & 18.32 & 19.34 & 23.3 & 19.96 & 20.78 & 21.85 & 21.45 \\
    Our Few TensoRF & \textcolor{blue}{25.68} & \textcolor{blue}{27.70} & \textcolor{blue}{18.88} & \textcolor{blue}{21.09} & 24.56 & 21.64 & \textcolor{blue}{22.04} & 27.99 & 23.70 \\
    Our Few TensoRF (Fine tune) & \textcolor{red}{25.73} & \textcolor{red}{27.71} & 18.04 & \textcolor{red}{21.36} & \textcolor{red}{27.44} & 23.77 & \textcolor{red}{22.36} & \textcolor{red}{29.71} & \textcolor{red}{24.52} \\
    \bottomrule
  \end{tabular}
  \label{tab:table1}
\end{table}

\begin{table}[htb!]
 \caption{Training time comparison on the Hotdog scene. The bold markings indicate the most optimal results obtained in our experiments.}
  \centering
  \begin{tabular}{lll}
    \toprule    
    \textbf{Metrics} & \textbf{PSNR\(\uparrow\)} & \textbf{Training time\(\downarrow\)} \\    
    \midrule
    FreeNeRF\_50k (repro.) & 29.40 & 04:55:00 \\
    FreeNeRF\_15k (repro.) & 28.02 & 01:25:40 \\
    TensoRF (repro.) & 21.85 & 00:15:00 \\
    Our Few TensoRF & 27.99 & 00:15:20 \\
    Our Few TensoRF (Fine tune) & \textbf{29.71} & \textbf{00:10:22} \\
    \bottomrule
  \end{tabular}
  \label{tab:table2}
\end{table}

\begin{figure}[htb!] 
    \centering
    \includegraphics[width=1.0\textwidth]{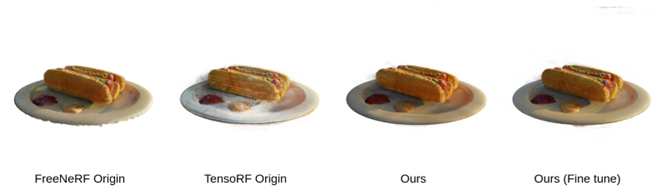}
    \caption{\textbf{Comparing the view rendered of Hotdog object between methods.}}
    \label{fig:fig5}
\end{figure}

To assess the training efficiency among methods, we adjusted FreeNeRF iterations to 15000 to establish a common reference. Lower training times and higher PSNR values are preferable. After numerous experiments, we observed minimal differences in training times across scenes, typically varying by only 2-3 minutes for each method. Therefore, in Table \ref{tab:table2}, we opted to reproduce FreeNeRF 15000 iterations on HotDog alone to save time and Fig. \ref{fig:fig5} shows the result of four methods for Hotdog object. Additionally, as anticipated, FreeNeRF 15000 iterations did not optimize well enough to match the performance of FreeNeRF\_50k. Our fine-tuned model demonstrated superior efficiency and rendering quality compared to other methods. In summary, Our Few TensorRF outshines FreeNeRF and TensorRF in average PSNR, with finetuning enhancing performance across scenes. Despite the intricacies of the challenging Drum scene, the proposed
model demonstrates superior efficiency and rendering quality. The findings highlight the effectiveness of our Few TensorRF methods in addressing few-shot challenges, making it a promising approach for highquality 3D reconstruction.

\begin{figure}[htb!] 
    \centering
    \includegraphics[width=1.0\textwidth]{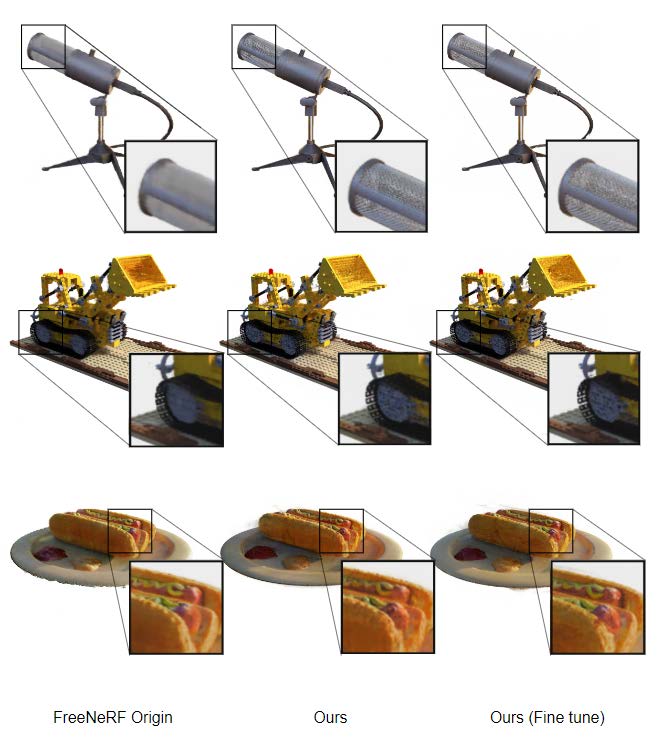}
    \caption{\textbf{Visual comparison between methods.}}
    \label{fig:fig8}
\end{figure}

Furthermore, Fig. \ref{fig:fig8} illustrates the underlying reasons for the superior PSNR scores achieved by the proposed methods in these scenes. While the FreeNeRF method tends to blur fine details within the scene, our approach provides more precise refinement across nearly all scenarios evaluated in Synthesis NeRF.

\begin{figure}[htb!] 
    \centering
    \includegraphics[width=1.0\textwidth]{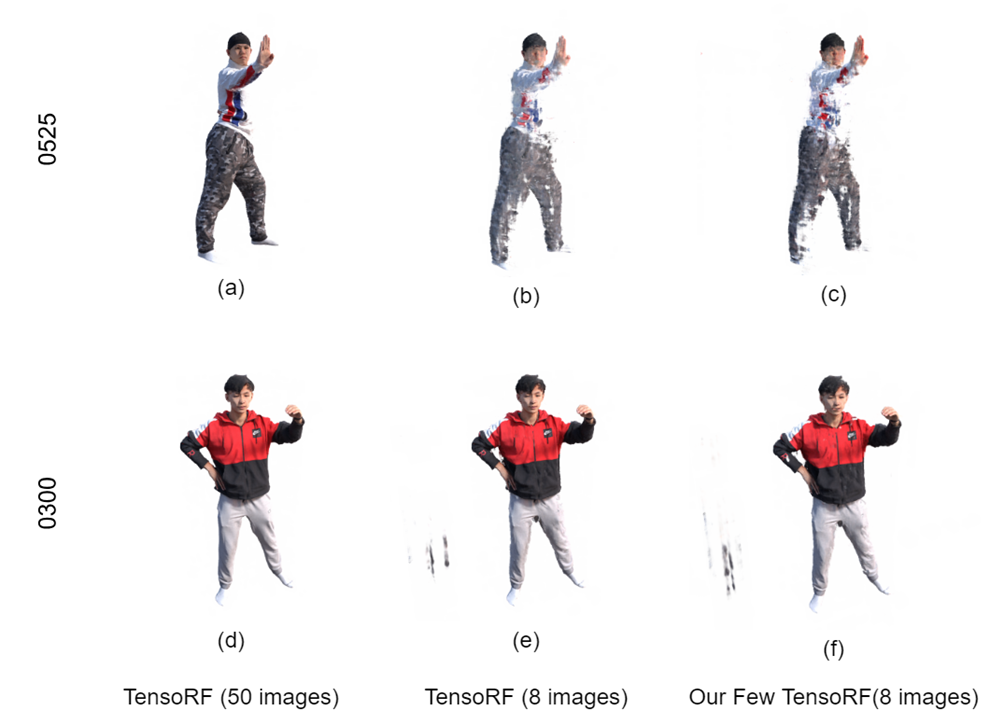}
    \caption{\textbf{Comparing the quality of rendered images with different numbers of images between TensoRF and Few TensoRF.}}
    \label{fig:fig6}
\end{figure}

\textit{Performance on THuman 2.0 dataset}. Figure \ref{fig:fig6} presents a comparison between two objects, 0300 and 0525, from the THuman0.2 dataset. For this analysis, models were trained on either 50 or 8 images using TensoRF, while our Few TensoRF was implemented with a visible frequency ratio of 0.8. All referenced models utilized VM decomposition and were trained for an identical number of iterations (15,000).

Table \ref{tab:table3} offers a detailed summary of the performance of three methods on the THuman 2.0 dataset, with particular emphasis on objects 0525 and 0300. As expected, the original TensoRF models—trained with a larger number of input images—exhibited superior image rendering performance, providing high accuracy and intricate detail attributable to increased data availability. To assess few-shot learning capability, we reduced the number of training views to eight, paralleling the methodology used in the Synthesis Nerf experiment. The results indicate that the Few TensoRF models show diminished performance relative to the original TensoRF.

\begin{figure}[htb!] 
    \centering
    \includegraphics[width=0.7\textwidth]{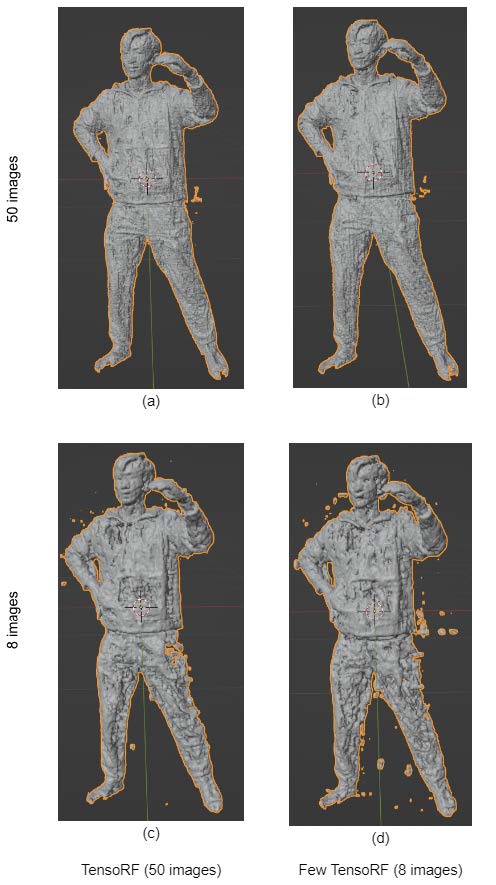}
    \caption{\textbf{Comparing 3D meshes rendered for 0300 object.}}
    \label{fig:fig7}
\end{figure}

The comparative 3D meshes displayed in Fig. \ref{fig:fig7} further corroborate the PSNR score findings reported in Table 3. Training TensoRF with 50 images enabled accurate reconstruction of object details such as clothing, facial features, fingers, and toes. Conversely, models trained with fewer images generated meshes with numerous holes and exhibited instability during object rendering. Notably, the Few TensoRF model trained on only eight images produced increased noise around the object compared to the original TensoRF model trained with more extensive data.

\begin{table}[htb!]
 \caption{Comparison of PSNR parameters between TensoRF and Few TensoRF with different numbers of input images.}
  \centering
  \begin{tabular}{lll}
    \toprule    
    \textbf{PSNR} & \textbf{0525} & \textbf{0300} \\    
    \midrule
    TensoRF (50 images) & 40.98 & 45.58 \\
    TensoRF(8 images) & 28,37 & 34.28 \\
    Few TensoRF (8 images) & 27.37 & 34.00 \\
    \bottomrule
  \end{tabular}
  \label{tab:table3}
\end{table}

\subsection{Discussion}
The results provide a positive outlook on the potential development of our method in future applications. The commendable performance on the Synthesis NeRF dataset positions our approach favorably among prior studies in novel view synthesis, establishing a benchmark for subsequent work on the THuman 2.0 dataset. However, despite the strengths exhibited, certain limitations are evident. Challenges emerged in specific scenes, notably the "Drums" scene in Synthesis NeRF, where our method encountered difficulties. The exact nature of this issue remains elusive, possibly attributable to the intricate details of the "Drums" scene.

Interestingly, experiments on the novel THuman 2.0 dataset yielded compelling results, showcasing higher PSNR metrics compared to standard NeRF datasets used in prior studies. Yet, the rendered images exhibited notable noise, highlighting an area for improvement in future endeavors. This underscores the need for a more robust approach to enhance the quality of 3D reconstruction meshes, particularly in the context of the THuman 2.0 dataset. Our method's distinctive strength lies in its ability to achieve efficiency across three pivotal scenarios: fast training time, utilization of fewer images, and maintaining an acceptable level of general quality. This emphasis on efficiency aligns with our primary goal of contributing to the advancement of NeRF-like techniques. However, evaluating the overall performance of Few TensoRF on THuman 2.0 dataset is challenging due to the limited experimentation on only two out of 500 objects in the dataset. Within the scope of this study, we focused on experimenting with our method on specific THuman 2.0 objects instead of training on the entire dataset for a comprehensive benchmark.

Looking ahead, our focus on THuman 2.0 opens avenues for refining methods and addressing the identified limitations. In a broader context, NeRF-like methods, including our Few-TensoRF, hold promise not only for reconstructing 3D models but also for excelling in novel view synthesis—a crucial technique poised to revolutionize the fields of virtual reality (VR) and augmented reality (AR) applications.

\section{Conclusion}
\label{sec:conclusion}
The proposed Few-TensoRF presents a step forward in the realm of novel view synthesis, showing promise across various scenarios. By addressing challenges and limitations, our approach stands as a valuable asset for real-world applications with resource constraints. The promising results achieved on both Synthesis NeRF and THuman 2.0 datasets, coupled with a focus on overcoming identified limitations, underscore the potential impact and versatility of our Few-TensoRF method in advancing the capabilities of 3D reconstruction and view synthesis technologies.

In conclusion, the Few TensoRF method represents a significant advance in 3D reconstruction, achieving faster training times and improved efficiency. Integrating FreeNeRF and TensoRF capabilities, the proposed Few TensoRF demonstrates versatility across diverse datasets, contributing to the broader field of 3D reconstruction and promising advancements in applications requiring faster training times and dynamic scene captures.

\section*{Declaration of conflicting interest}
The authors declared no potential conflicts of interest with respect to the research, authorship, and/or publication of this article.

\section*{Data availability statement}
Data supporting the findings of this article are publicly available at \url{https://www.kaggle.com/datasets/nguyenhung1903/nerf-synthetic-dataset} and \url{https://github.com/ytrock/THuman2.0-Dataset}.

\section*{ORCID iD}
Thanh-Hai Le: \url{https://orcid.org/0000-0002-3212-3940}

\bibliographystyle{unsrt}  
\bibliography{references}  

\end{document}